\documentclass{article}
\usepackage{arxiv}

\usepackage[utf8]{inputenc} %
\usepackage[T1]{fontenc}    %
\usepackage{hyperref}       %
\usepackage{url}            %
\usepackage{booktabs}       %
\usepackage{amsfonts}       %
\usepackage{nicefrac}       %
\usepackage{microtype}      %
\usepackage{lipsum}		%
\usepackage{graphicx}
\usepackage{natbib}
\usepackage{doi}

\usepackage{amsmath}
\usepackage{adjustbox}
\usepackage{booktabs}
\usepackage{framed,multirow}
\usepackage{xcolor}

\newcommand{\changed} [1] {\textcolor{black}{#1}}

\title{Transformer-based Image Generation from Scene Graphs}

\author{ \href{https://orcid.org/0000-0002-3906-797X}{\includegraphics[scale=0.06]{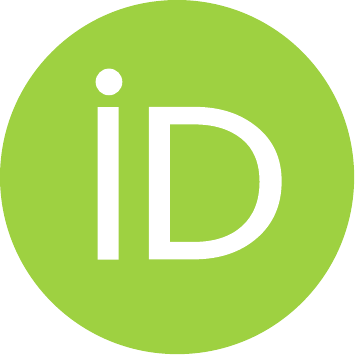}\hspace{1mm}Renato ~Sortino}\\
	PeRCeiVe Lab\\
	University of Catania\\
	Catania, Italy \\
	\texttt{renato.sortino@phd.unict.it} \\
	\And
	\href{https://orcid.org/0000-0002-2441-0982}{\includegraphics[scale=0.06]{orcid.pdf}\hspace{1mm}Simone Palazzo} \\
	PeRCeiVe Lab\\
	University of Catania\\
	Catania, Italy \\
	\texttt{simone.palazzo@unict.it} \\
	\And
	\href{https://orcid.org/0000-0001-6653-2577}{\includegraphics[scale=0.06]{orcid.pdf}\hspace{1mm}Concetto Spampinato} \\
	PeRCeiVe Lab\\
	University of Catania\\
	Catania, Italy \\
	\texttt{concetto.spampinato@unict.it} \\
}

\date{}

\renewcommand{\headeright}{}
\renewcommand{\undertitle}{}
\renewcommand{\shorttitle}{Transformer-based Image Generation from Scene Graphs}

\hypersetup{
pdftitle={Transformer-based Image Generation from Scene Graphs},
pdfsubject={q-bio.NC, q-bio.QM},
pdfauthor={Renato Sortino, Simone Palazzo, Concetto Spampinato},
pdfkeywords={scene graphs, transformers, generative models, conditional image generation},
}

\begin{document}
\maketitle

\begin{abstract}
	
Graph-structured scene descriptions can be efficiently used in generative models to control the composition of the generated image. Previous approaches are based on the combination of graph convolutional networks and adversarial methods for layout prediction and image generation, respectively. In this work, we show how employing multi-head attention to encode the graph information, as well as using a transformer-based model in the latent space for image generation can improve the quality of the sampled data, without the need to employ adversarial models with the subsequent advantage in terms of training stability. \\
The proposed approach, specifically, is entirely based on transformer architectures both for \emph{encoding} scene graphs into intermediate object layouts and for \emph{decoding} these layouts into images, passing through a lower dimensional space learned by a vector-quantized variational autoencoder.
 Our approach shows an improved image quality with respect to state-of-the-art methods as well as a higher degree of diversity among multiple generations from the same scene graph.  We evaluate our approach on three public datasets: Visual Genome, COCO, and CLEVR. We achieve an Inception Score of 13.7 and 12.8, and an FID of 52.3 and 60.3, on COCO and Visual Genome, respectively. We perform ablation studies on our contributions to assess the impact of each component.
Code is available at \url{https://github.com/perceivelab/trf-sg2im}

\end{abstract}

\keywords{scene graphs\and transformers\and generative models\and conditional image generation}

\section{Introduction}

\begin{figure*}
	\centering
	\includegraphics[width=0.75\textwidth]{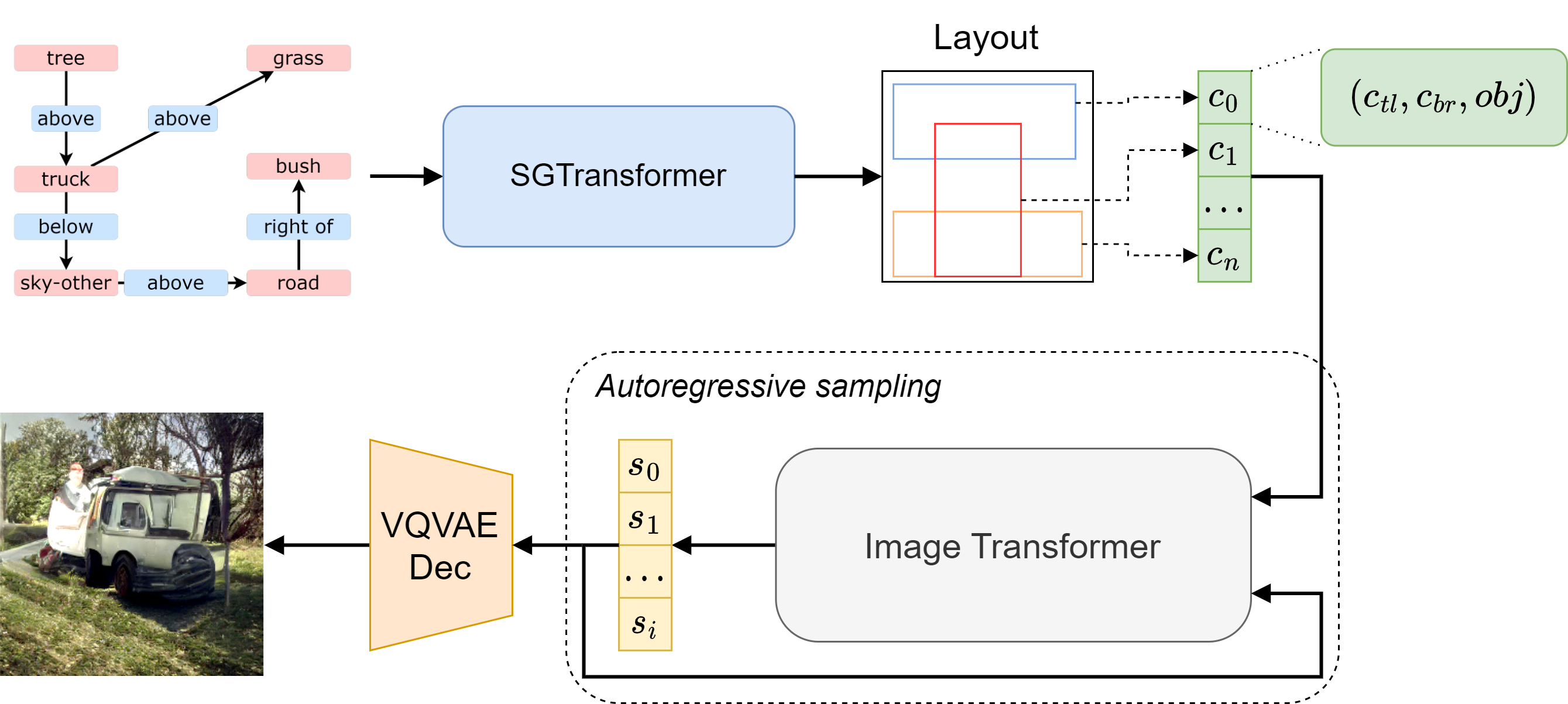}
	\caption{Overview of our framework. The scene graph is fed to our SGTransformer module, which updates the node feature vectors with multi-head attention on the neighboring nodes and the edge feature vectors to predict the layout. This is treated as a sequence of discrete elements $c_i$, which are triplets where the first two elements are bounding box coordinates, top left and bottom right, respectively, and the last one is the object category. The Image Transformer predicts a sequence of codebook indices $s_i$ that are then reshaped and projected back onto the pixel space using a pre-trained VQVAE decoder.}
	\label{fig:arch}
\end{figure*}

Conditional image generation has recently shown promising results in generating high-resolution images, especially in recent formulations involving text-to-image approaches with transformers~\citep{dalle,clip-vqgan} or diffusion models~\citep{dalle2,imagen,ldm}. 
The evolution of the generative model architectures has significantly contributed to make conditional image generation possible. Generative Adversarial Networks (GAN)~\citep{gan} have quickly evolved in less than a decade, reaching high sample quality through architectural improvements~\citep{stylegan2,cyclegan,discogan} and conditioning information~\citep{cgan,text2image,pix2pix}. Recent approaches, particularly the ones based on transformers~\citep{transformer} and diffusion models~\citep{ddpm}, achieve even better results and avoid the training instability typical of adversarial approaches, but at the cost of an increase in computing and data requirements. \\
While generating data from a random noise distribution, possibly partitioned for conditional generation, can have interesting applications, using human-readable input to control the generation process makes these approaches more suitable for addressing specific problems by limiting the generated samples to a particular sub-domain.
Text, in particular, is a flexible modality that allows unlimited combinations of conditioning inputs, including complex and abstract concepts, and can potentially lead to the generation of high-quality and high-quantity images~\citep{stable-diffusion}.
However, using text as a conditioning signal comes with several drawbacks: natural language sentences may be long and loosely structured; semantics depend strongly on syntax and language; it inherently contains ambiguity as many different sentences can express the same concept, which can lead to unstable training, especially with low model capacity. 
Thus, in contexts where description constraints are important --- for instance, highly-specialized domains (e.g., medicine, astronomy, agriculture) --- text representations of a specific scene might not suffice. \emph{Scene graphs}~\citep{sg2im} offer a valid alternative to structure the description of visual data using graphical syntax, where nodes represent objects in the scene and edges model the spatial/semantic relationships between them. In particular, scene graphs provide a comprehensive and flexible way of describing objects and their relations in a scene. Nevertheless, 
a major challenge in employing scene graphs to control image generation lies in the complexity of matching visual and graph features. A common effective strategy that has been employed to overcome this limitation is to estimate, from the graph, an intermediate scene layout representation that, in turn, is used to guide the generation process~\citep{sg2im,pastegan}. 
However, the choice of using adversarial models for the image generation process, adopted by most existing approaches, may limit the potentiality of this conditional generation, as adversarial-based methods are characterized by unstable training. \\
To overcome this limitation, we propose a fully transformer-based approach for scene-graph-to-image generation. More specifically, we exploit the generalization capabilities of transformers in processing arbitrary graphs through multi-head attention mechanisms and in autoregressively composing long sequences. The devised model is shown in Fig.~\ref{fig:arch}: we first predict a scene layout from scene graphs through the \emph{SGTransformer} module, that computes attention on neighboring nodes of a graph and the respective edge features, and predicts the layout, a set of bounding boxes and corresponding labels. 
We then map the layout information to a discrete space, by aggregating the coordinates of each box and the respective object labels into triplets, and use this information to condition another transformer, \emph{Image Transformer (ImT)} which learns the joint distribution between images and layouts. Generation in the Image Transformer is posed as a classification task on a sequence of image tokens and leverages the transformer decoder's capability to generate long sequences without losing context information. The sequence of predicted tokens is finally translated into an image through a pre-trained vector quantized autoencoder, i.e., VQVAE~\citep{vqvae}, that learns to project data from the pixel space $\mathcal{X}$ to a latent space $\mathcal{Z}$ of quantized feature vectors, and vice versa.\\ 
We show the effectiveness and flexibility of our scene-graph-to-image approach on standard benchmarks such as Visual Genome (VG) and COCO, where we report visual examples showing its robustness to (both minor and major) perturbations in the input scene graph, while ensuring significant variability and diversity in the synthesized images. 
We also compare our method with current state-of-the-art approaches, yielding generally better qualitative and quantitative results in terms of both generated layouts and images. We also carry out extensive ablation studies that substantiate our design choices. 
We finally conclude by concretely showing how the strategy to explicitly generate intermediate layout representations yields better performance than carrying out cross-attention between encoder and decoder modules of the transformer, thus substantiating again our choice.  \\

\label{sec:intro}

\section{Related Work}

\textbf{Transformers as generative models.} Transformers~\citep{transformer} have shown to perform better than convolutional approaches in several computer vision tasks, such as object detection~\citep{detr,yolos} and classification~\citep{vit,deit}. Vision transformers have been employed for generative models as well, trained both in an adversarial~\citep{transgan,ganformer,ganformer2} and a supervised way~\citep{vqgan, dalle}. TransGAN~\citep{transgan} employs two transformer encoders that are trained adversarially and act as generator and discriminator in traditional GANs. GANformer~\citep{ganformer} introduces a bipartite attention mechanism between the image and a latent noise vector used to start the generation, which helps to compute attention between latent features and image features and couple parts of the latent vector with image regions. GANformer2~\citep{ganformer2} evolves this work by adding conditioning information to control the generation process. 
Transformers are better than CNNs at modeling long-range sequences, which makes them suitable for autoregressive generation tasks. In particular, if images can be represented as a sequence of integers, using codebooks of quantized image features, transformers can be efficiently used for generating images at high resolution, as shown in VQGAN~\citep{vqgan}. This method has been improved to perform also text-to-image synthesis by integrating CLIP~\citep{clip} and using it to guide the model towards generating images that match a text description.

\textbf{Scene graph to image.} Research related to the task of image generation conditioned on scene graphs has seen interesting developments since the launch of the pioneering work of SG2IM~\citep{sg2im}. This work demonstrated how image generation can effectively be conditioned on scene graphs with good results, employing a Graph Convolutional Network (GCN) to generate a scene layout, and then a Generative Adversarial Network conditioned on this layout to generate the image. Since SG2IM, other approaches improved this task in several ways. PasteGAN~\citep{pastegan} enhances the GAN by conditioning it on image crops that contain single objects corresponding to scene graph nodes. The work proposed by~\citep{contextsg2im} builds on SG2IM by adding context information to the image generator. CanonicalSG2IM~\citep{canonicalsg2im} adds transitive and converse relationships with a certain probability, to enrich the graph with additional information. Other methods use different approaches to improve the generation process, for instance by processing \emph{subject-predicate-object} relations individually~\citep{vo2020visualrelation}, by posing the task as a meta-learning problem~\citep{migs}, or by iteratively modifying the scene graph while retaining previously-generated content~\citep{mittal2019interactive}.

All of these works employ a two-stage process to generate the image. Most of them generate a scene layout first, and then use it to condition a GAN generator. Our work follows the same two-stage pipeline, but adopts a fully transformer-based architecture, trained in a standard supervised way, thus avoiding the instability and mode collapse problems of adversarial training. This allow us to synthesize higher-quality samples compared to existing models, while increasing model flexibility, sensitivity to small graph changes, and diversity of generated samples.

\label{sec:related}

\section{Method}
In this section, we discuss in detail the architectural elements that compose our framework. Our goal is to exploit the high capacity of transformer-based architectures to extract meaningful information from the scene graph, with a twofold objective: 1) to produce a plausible and coherent layout of the scene using information from the Laplacian matrix and the graph's edge features; and 2) to exploit the capability of modeling long sequences to generate the output image as a sequence of codebook indices. 
We thus propose a two-stage approach, which first generates a layout from the scene graph and then conditions the image generation process on the predicted layout.

\subsection{Scene graph to layout}
\begin{figure}[!ht]
	\centering
	\includegraphics[width=0.4\textwidth]{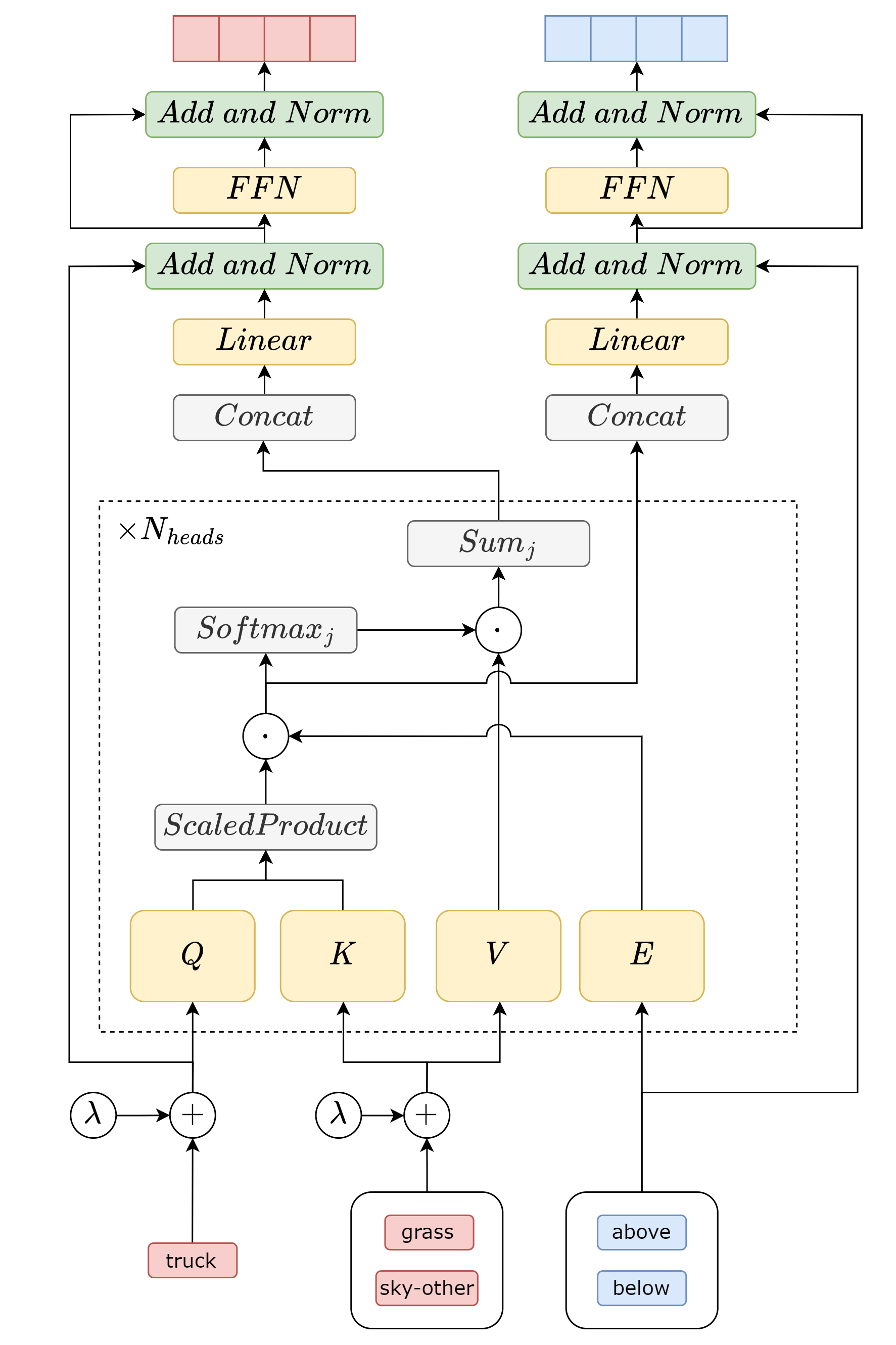}
	\caption{Detail of a SGTransformer layer. Each node is used as query, while its neighboring nodes $j$ are mapped to key-value pairs. Edge features are multiplied before the softmax operations with the product of $Q$ and $K$, and the output of this operation is used to update the edge features.}
	\label{fig:sgtransformer}
\end{figure}

Scene graphs describe the composition of an image by defining the relationships between its objects in the form of a directed graph $G \left( V,E \right)$. Objects are represented as graph nodes containing the object category $o \in \left\{0, \dots, N_o\right\}$, while edges encode the type of relationship $p \in \left\{0, \dots, N_p\right\}$ that occurs between two nodes, with $N_o$ and $N_p$ being the number of object categories and relationship types, respectively. Nodes and edges are projected into a higher-dimensional space, thus each node $i$ is characterized by a feature vector $h_i \in \mathbb{R}^{d_n}$; for each node $j$ in its neighborhood $\mathcal{N}_i$, the corresponding edge is described by a feature vector $e_{ij} \in \mathbb{R}^{d_e}$. 
State-of-the-art scene-graph-to-image approaches use Graph Convolutional Networks (GCNs) to encode the graph representation by updating each node's feature vector with information from neighboring nodes and edges. \citet{graphtransformer} point out how GCNs, and particularly Graph Attention Networks~\citep{gat}, can be interpreted as a generalized case of transformers, where the attention mechanism is computed only on neighboring nodes. Motivated by the high versatility of transformers in this context, we employ such an architecture to encode our scene graph. A first challenge in this formulation lies in the definition of a \emph{positional encoding} that preserves the structure of the input. Typically, the positional encoding helps to maintain the relative ordering of the elements within the sequence. When dealing with graphs, the concept of ``ordering'' of nodes has no meaning, as graphs are unordered by definition. However, the Laplacian matrix provides geometric information on the graph, and~\citet{graphtransformer} show that it can be used as a positional encoding to enrich the encoded graph representation. We employ the architecture proposed by~\citet{graphtransformer} as our scene graph encoder, which we call \emph{SGTransformer}. An overview of this module's architecture is shown in Fig.~\ref{fig:sgtransformer}.

The graph Laplacian $\Delta \in \mathbb{R}^{n \times n}$ is defined, in its symmetrically normalized form, in the following way:
\begin{equation}
    \Delta = I - D^{-1/2}AD^{-1/2}
\end{equation}
where $D \in \mathbb{R}^{n \times n}$ represents the degree matrix and $A \in \mathbb{R}^{n \times n}$ is the adjacency matrix. As a positional encoding, we use the $l$ smallest non-trivial eigenvectors of $\Delta$, which we indicate with $\lambda$, and randomly flip their sign during training, due to their multiplicity given by arbitrary sign. These eigenvectors are projected, via a learnable matrix, onto a space with the same dimensionality of the node embeddings and summed to them.

The SGTransformer architecture is composed of a stack of blocks, each of which computes multi-head attention between each node $h_i$ and its neighbors $h_{j} \in \mathcal{N}_{i}$. This attention mechanism can be interpreted as a masked self-attention where, for each node, all the non-neighboring nodes are masked out. This reduces the complexity of the attention operation from $O(n^2)$ to $O(nm)$ where $n$ is the number of nodes and $m$ the maximum number of edges per node. In addition, when computing attention, we take into account the edge features $e_{ij}$, as they contain important information to describe the scene structure.

For each node $h_i$, the multi-head attention mechanism is defined as follows:

\begin{equation}
    \text{Attention}\left( Q, K, V, E \right) = \text{softmax}\left(\frac{h_iW^Q \cdot h_jW^K}{\sqrt{d_k}} \right) \cdot h_jW^V \cdot e_{ij}W^E.
    \label{eq:attn}
\end{equation}
$W^Q_k$, $W^K_k$, $W^V_k$, $W^E_k  \in \mathbb{R}^{d_k \times d}$ are learnable projection matrices, $k \in \left\{ 0, \dots, N_{heads} \right\}$, and $d_k = d / N_{heads}$.
Details on the layer of the SGTransformer are shown in Figure~\ref{fig:sgtransformer}.

The output of SGTransformer is then passed to two MLP heads that predict the bounding box and class label for each node vector. The set of bounding boxes and labels for a graph defines a layout.
We train the SGTransformer module with a compounded layout loss $\mathcal{L}_{layout}$, defined as follows:

\begin{equation}
    \mathcal{L}_{layout} = \mathcal{L}_{box} + \mathcal{L}_{label} + \mathcal{L}_{iou},
\end{equation}
where $\mathcal{L}_{box}$ is a $L_2$ loss on the bounding box coordinates, $\mathcal{L}_{label}$ is a cross-entropy on the labels, and $\mathcal{L}_{iou}$ is a distance IoU loss~\citep{diou} on the bounding boxes.

\subsection{Images as sequences of discrete tokens}
There are several methods that perform image generation conditioned on a layout~\citep{lostgan,layout2im2,layout2im3} and some of them have been employed as the second stage of scene-graph-to-image approaches. State-of-the-art approaches~\citep{sg2im,canonicalsg2im,pastegan} rely on adversarially trained networks for image generation. We propose a transformer-based image generator, which we refer to as \emph{Image Transformer (ImT)}. The goal is to use the affirmed capability of transformers to efficiently process long sequences, widely explored in NLP, for image generation. To accomplish this, we need to express image data as a sequence of tokens in a similar way as is done for words in a sentence. Vector-Quantized Variational Autoencoders (VQVAE)~\citep{vqvae} encode the input RGB image $x \in \mathbb{R}^{H \times W \times 3}$ into a latent representation $\hat{z} \in \mathbb{R}^{h \times w \times n_z}$, where $h$ and $w$ are the number of feature vectors along the $H$ and $W$ sizes of the input image, and $n_z$ represents each vector's size. Each vector is then used to retrieve the nearest feature vector $z_q$ from a codebook:
\begin{equation}
    z_{\mathbf{q}}= \left(\underset{z_k \in \mathcal{Z}}{\arg \min }\left\|\hat{z}_{i}-z_k\right\|\right) \in \mathbb{R}^{N \times n_z}
\end{equation}
Here, $z_k$ are the codebook vectors and $z_i$ the i-th feature vector of the encoded input, with $i \in \left\{0,\dots, hw\right\}$.
The VQVAE is trained to reconstruct $\hat{x}$ from input data $x$ while defining a meaningful codebook using a reconstruction loss $\mathcal{L}_r$, a quantization loss $\mathcal{L}_q$, and a commitment loss $\mathcal{L}_c$, defined as follows:

\begin{equation}
    \centering
    \mathcal{L}_{vqvae}(x) = \mathcal{L}_{r}(x) + \mathcal{L}_{q}(x) + \mathcal{L}_{c}(x)
\end{equation}
\begin{equation}
     \mathcal{L}_{r}(x) = \left\| x - \hat{x} \right\| 
\end{equation}
\begin{equation}
    \mathcal{L}_{q}(x) = \left\|\text{sg} \left[\mathcal{E}(x)\right] - z_k\right\|_2^2
\end{equation}
\begin{equation}
    \mathcal{L}_{c}(x) = \left\|\mathcal{E}(x) - \operatorname{sg}[z_k]\right\|_2^2
\end{equation}

Here, $\text{sg}$ refers to the stop-gradient operator. The quantization loss $\mathcal{L}_q$ is used to minimize the distance between the encoded vectors and the codebook vectors, while the commitment loss $\mathcal{L}_c$ is employed as a form of regularization of the latent space to prevent it from growing indefinitely. 

The quantization operation causes high-frequency information loss in the reconstructed image. Therefore, in addition to these losses, we add a patch discriminator loss $\mathcal{L}_{disc}$ and a perceptual loss $\mathcal{L}_{VGG}$, as suggested by~\citet{vqgan}.

\subsection{Layout to Image}
Once we learned the feature codebook, we can use the VQVAE to efficiently encode the images from the pixel space $\mathcal{X}$ to the latent space $\mathcal{Z}$ and vice versa. To generate new samples, we need to estimate the prior distribution of the latent codebook vectors. The Image Transformer is an autoregressive transformer that models the conditional distribution $p(s_i~|~s_{<i})$, where $s$ represents the codebook indices (image tokens) and $i$ is the step within the iterative process. It follows the architecture of GPT~\citep{gpt}, employing multiple masked self-attention blocks to process the input sequence. The transformer attends only the previous blocks in the sequence by masking out the next tokens in the sequence, $\left\{ s_j \right\}_{j>i}$. 
We condition the generation on the layout predicted from the SGTransformer. To this purpose, we encode the layout by mapping each object into a triplet of the form $o = (c_y, tl_y, br_y)$, as done by~\citep{layout-vqgan}. Here, $c_y$ represents the class index, $tl_y$ and $br_y$ the bounding box coordinates on a diagonal, top left and bottom right, respectively. The sequence of encoded layout objects $c \in \left\{ 0, \dots, N_o\right\}$ is prepended to the sequence of image tokens, and self-attention is computed on the entire sequence, combining information from the layout and the generated indices. Thus, the transformer learns to model the joint distribution of the condition $c$ and the sequence of tokens $s$: $p(s_i~|~s_{<i}, c)$.

The Image Transformer learns this prior by optimizing the following log-likelihood:
\begin{equation}
    \mathcal{L}_{\text {Transformer }}=\mathbb{E}_{x, c \sim p(x, c)} \left[ -\log p(s~|~c) \right]
\end{equation}

\label{sec:method}

\section{Experimental Results}
We hereby present the experimental protocol used to evaluate our model's performance and compare it to the state of the art, describing in details the employed datasets, the training procedure and our results, both quantitatively and qualitatively.

\subsection{Dataset}
For comparability with state-of-the-art methods, we trained and evaluated our model on three public image datasets that provide annotations in the form of object-to-object relations, which can be expressed as scene graphs: Visual Genome~\citep{vg}, COCO-Stuff~\citep{coco-stuff}, and CLEVR-Dialog~\citep{clevr-dialog}.

Visual Genome (VG) contains 108,077 images, with annotations in JSON format specifying bounding boxes, object categories, attributes, and relationships. We filter the annotations and generate the scene graphs as done by~\citep{sg2im}, keeping object and relationship instances that occur at least 2000 and 500 times respectively. This results in 178 objects categories and 45 relationship types, describing actions and spatial relationships.
The COCO-Stuff dataset extends COCO~\citep{coco} by adding 91 object categories with bounding box annotations. This augmented dataset has a total of 45,000 images and 171 object categories. Scene graphs are generated following~\citep{sg2im}, computing the relative positions of a subset of all the possible bounding box pairs and inferring the spatial relation among the contained objects and generating a total of 6 relationship types: \emph{above}, \emph{below}, \emph{left of}, \emph{right of}, \emph{inside}, and \emph{surrounding}.
For VG and COCO-Stuff, we use the same train/test split defined by~\citep{sg2im}
CLEVR-Dialog~\citep{clevr-dialog} is a procedurally generated dataset with associated scene descriptions, expressed as question-answers pairs describing the scene. It includes a total of 70,000 images for the training set and 10,000 in the validation set. Scene graphs from CLEVR Dialog have been generated following the procedure shown in~\citep{canonicalsg2im}. 
We normalize to zero mean and unitary standard deviation (per channel) and resize the images to 128$\times$128 for all datasets.

\subsection{Training procedure}
We hereby describe the training procedure we followed to lead our framework to convergence.

The latent codebook contains $K=8192$ vectors, each of which has a dimensionality $n_z$ of 256. The VQVAE encoder compresses each image with a compression factor of $f = 8$, meaning that the input image of size $H \times W$ is compressed into a feature map of dimension $h \times w$, where $h = \frac{H}{f}$ and $w = \frac{W}{f}$. We chose this value based on the analysis of the compression factors reported by~\citep{ldm}, as it shows the best compromise between compression rate and information loss.

The SGTransformer is composed of 12 transformer layers, each with 12 attention heads, and an embedding size of 768. For the Laplacian positional encoding, we use the first $l = 8$ non-trivial eigenvectors and, as the number of nodes $N_o$ in a scene graph is variable and we have at most $N_o$ eigenvectors, in the case where $N_o < l$, we pad the eigenvector matrix with $l - N_o$ zero vectors.

\begin{figure}
	\centering
	\includegraphics[width=0.25\textwidth]{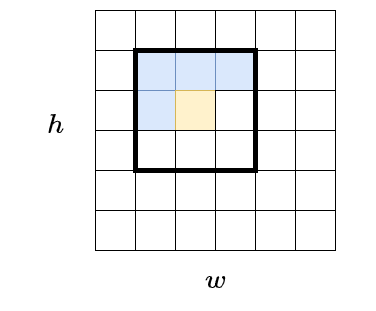}
	\caption{Representation of the decoder self-attention mask on the latent space. Elements in blue are attended by the element in yellow.}
	\label{fig:attn_mask}
\end{figure}

The Image Transformer is a stack of 40 layers, with an embedding size of 1408 and 16 attention heads.
Image data shows a higher correlation between neighboring pixels with respect to further ones. As the VQVAE is CNN-based, we maintain this correlation in the latent space. To exploit such a correlation, we employ a convolutional attention mask, as shown in Fig.~\ref{fig:attn_mask}, which attends only the previous elements of the sequence that fall within a 2D kernel. We use a kernel of size 7.
Additional high-level information on scene objects is provided by the layout encoding obtained from the SGTransformer output.

We train our model for 300 epochs with a learning rate of $10^{-4}$, using a One Cycle learning rate scheduler~\citep{onecyclelr} and the Adam optimizer~\citep{adam}.   

\subsection{Results}

We evaluate our approach on the following tasks: 1) \emph{scene graph to layout}, where we evaluate the performance of our SGTransformer module and compare with state-of-the-art models; 2) \emph{layout to image}, evaluating the quality of the generated images feeding the decoder with layouts predicted by the SGTransformer and ground-truth ones; 3) \emph{scene graph to image}, assessing the performance of the model when tested end-to-end.
We compare our approach with the following state-of-the-art methods: SG2IM~\citep{sg2im}, which is the first approach performing image generation from scene graph; CanonicalSG2IM~\citep{canonicalsg2im}, which enhances graph relationships by learning converse and transitive relationships; and PasteGAN~\citep{pastegan}

\begin{figure*}[!ht]
    \centering
    \includegraphics[width=0.95\textwidth]{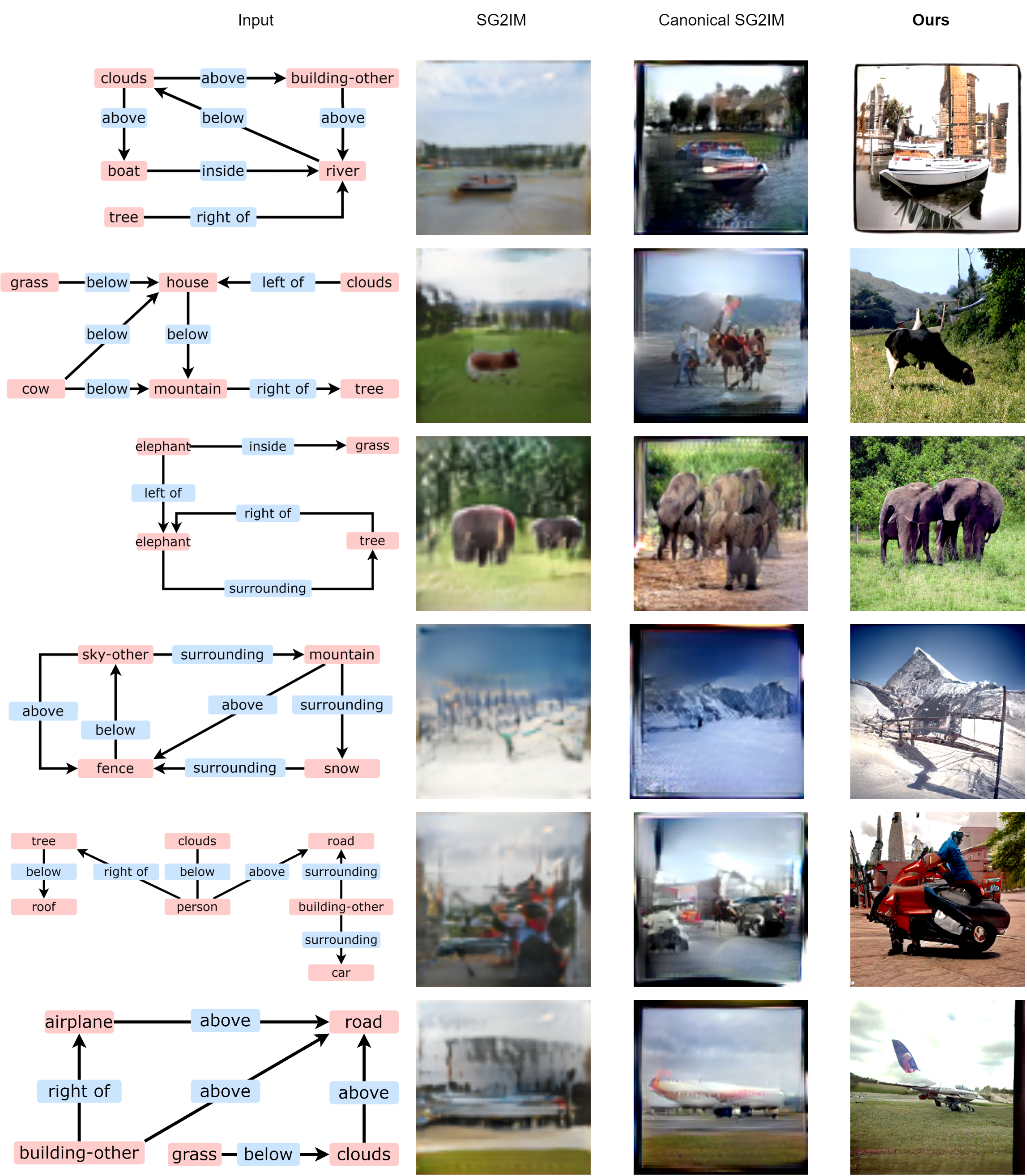}
    \caption{Visual comparison between our approach and SG2IM \citep{sg2im} and CanonicalSG2IM \citep{canonicalsg2im} on the COCO dataset}
    \label{fig:vis_comp}
\end{figure*}

\begin{figure}[!ht]
    \centering
    \includegraphics[width=0.48\textwidth]{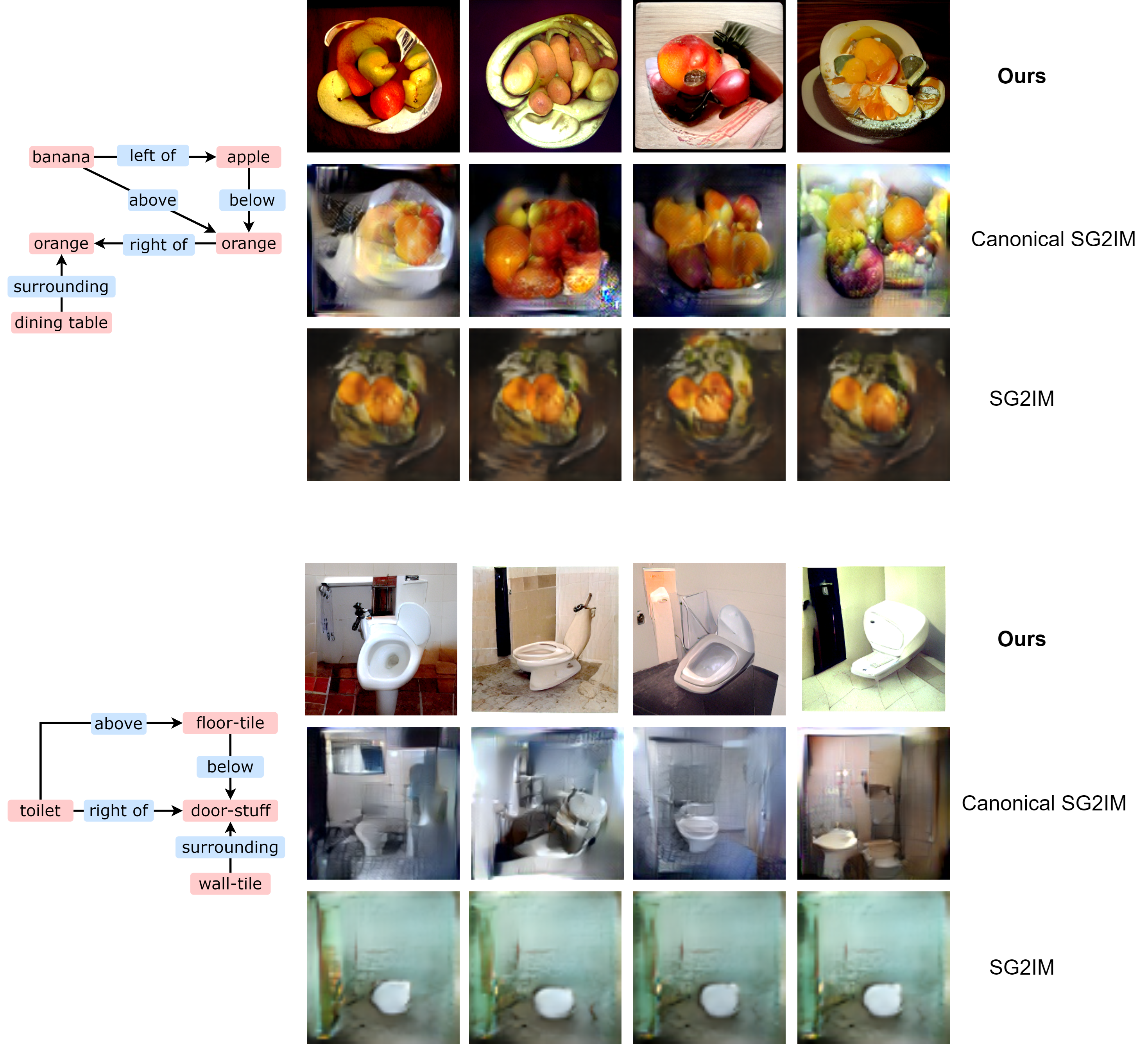}
    \caption{Diversity of several images sampled by the Image Transformer using the same layout. We compare our method with~\citep{canonicalsg2im} and ~\citep{sg2im}}
    \label{fig:diversity}
\end{figure}

We use FID~\citep{fid} and Inception Score (IS)~\citep{is} to compare the performance of our model with the state-of-the-art methods that employ intermediate layout generation for controlling image generation.
In Table~\ref{tab:layout2im} we report a comparison, in terms of image generation performance, with state-of-the-art models evaluated on Visual Genome and COCO. We computed FID and IS when image generation is controlled either by the intermediate generated layout or by the ground-truth layout (indicated with $^*$ in Table~\ref{tab:layout2im}).
\changed{On COCO, our method outperforms previous approaches in terms of both FID and IS, while on VG it reaches state-of-the-art performance in terms of Inception Score, but not FID.} This can be explained by the lower capacity of the VQVAE which, compressing the image, cuts high-frequencies in the output images, thus yielding a higher FID score.
A better Inception Score, instead, suggests a higher visual appearance diversity among the generated images, which translates into a higher capacity of our model to model the joint graph/image distribution.
These considerations can be also appreciated when comparing qualitatively our model against existing methods.
Fig~\ref{fig:vis_comp} shows how our model yields higher fidelity in the image details with respect to state-of-the-art methods. The VQVAE, trained using a perceptual loss and a discriminator loss, contributes to retrieve high-frequency components, resulting in a sharper image.
Fig.~\ref{fig:diversity}, instead, shows the capability of our \textit{Image Transformer} in synthesizing different images, given the same layout. The generated samples show a significant high degree of diversity among them (varying in color, lighting and object shapes), while being coherent with the input layout. Its capability to synthesize diverse samples, given the same scene graph, is even more evident when comparing the generated samples with those of existing methods.

\begin{figure*}
    \centering
    \includegraphics[width=0.95\textwidth]{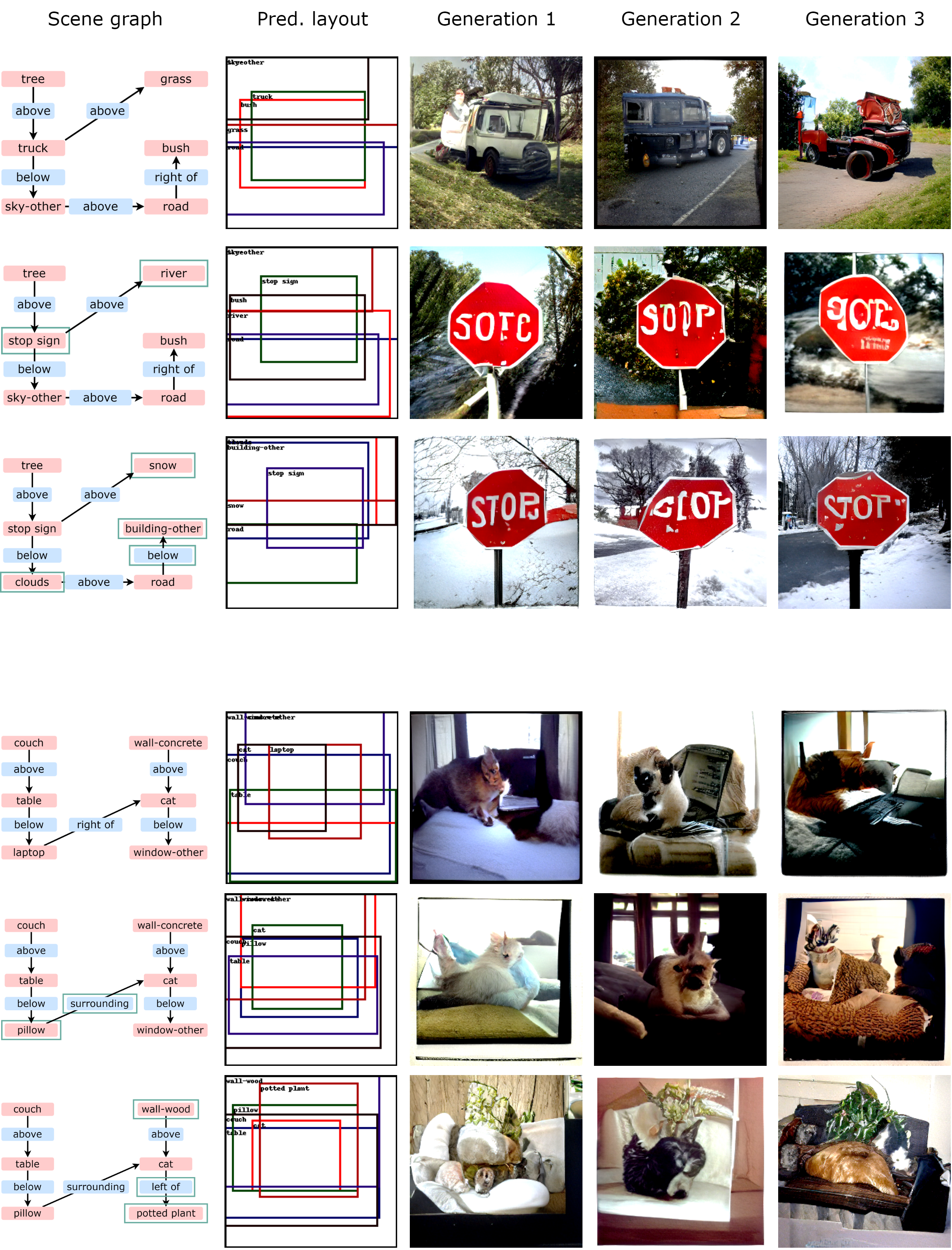}
    \caption{Manipulation of input scene graphs on the COCO-Stuff dataset. For each block of images: we start from an initial scene graph (top line) and as we move to the bottom, at each line, we modify some concepts/relations (highlighted in green) in the graph.}
    \label{fig:sg_changes}
\end{figure*}

We also assess qualitatively how the proposed approach is able to cope with perturbations in the input scene graph. Examples are reported in Fig.~\ref{fig:sg_changes}: from the first input graph we replaced "truck" with the "stop sign" and "grass" with "river", and the model inferred correctly to replace the truck in the scene with a stop sign in front of a river. These examples show the flexibility by our model in handling graph changes.

\begin{table}[!ht]
    \caption{Results in terms of image quality evaluated on COCO-Stuff and Visual Genome datasets. * refers to using ground-truth layouts as input to the Image Transformer. In bold best performance, in italic second best performance.}
    \centering
    \begin{adjustbox}{max width=\textwidth}
    \begin{tabular}{lcccc}
    \toprule
        {Method} & \multicolumn{2}{c}{Inception Score ($\uparrow$)} & \multicolumn{2}{c}{FID 
 ($\downarrow$)} \\
        \cmidrule(lr){2-3} \cmidrule(lr){4-5} 
         & COCO & VG & COCO & VG \\ 
        \midrule
        Real images & 23.0  $\pm$  0.4 & 22.8  $\pm$  1.7 & - & - \\ %
        \midrule
        SG2IM & 9.4 $\pm$ 0.2 & 8.3 $\pm$ 0.3 & 81.6 & 73.4 \\
        CanonicalSG2IM & 10.8 $\pm$ 0.5 & \textit{10.0 $\pm$ 0.7} & 73.8 & \textbf{46.4} \\
        PasteGAN & \textit{12.5 $\pm$ 0.8} & 8.5 $\pm$ 0.9 & \textit{70.5} & 62.8 \\
        Ours & \textbf{13.7 $\pm$ 0.9} & \textbf{12.8 $\pm$ 0.7} & \textbf{52.3 }& \textit{60.3} \\
        \midrule
        SG2IM* & 11.4 $\pm$ 0.5 & 10.8 $\pm$ 0.4 & 74.2 & 62.1 \\
        CanonicalSG2IM* & \textit{15.6 $\pm$ 0.5} & \textit{11.7  $\pm$  0.8} & \textit{54.7} & \textbf{36.4} \\
        PasteGAN* & 14.3 $\pm$ 0.7 & 10.9 $\pm$ 0.6 & 59.7 & 50.9 \\
        Ours* & \textbf{17.3 $\pm$ 0.9} & \textbf{15.5 $\pm$ 0.6} & \textbf{44.6} & \textit{42.6} \\
    \bottomrule
    \end{tabular}
    \end{adjustbox}
    \label{tab:layout2im}
\end{table}

We further evaluate the capability of our \emph{SGTransformer} to generate meaningful layouts (which then control the image generation process). 
Thus, we measure the mean Intersection over Union (mIoU) between the predicted layout and the ground truth layout associated to the input scene graph (available in the COCO and VG datasets). 
Table~\ref{tab:layout-gen} reports the comparison between our model and state-of-the-art approaches in terms of accuracy on the generate layout, showing that the SGTransformer outperforms state-of-the-art models on the scene-graph-to-layout task. This can be explained with a better capabilities of SGTransformer in learning the geometry of the input scene graph to better encode scene information, resulting in a more coherent layout, as also shown in Fig.~\ref{fig:sg_changes}.

\begin{table}[!ht]
    \centering
    \caption{Mean IoU on the bounding boxes predicted from the scene graph.}
        \begin{tabular}{lrr}
            \toprule
            {\textbf{Model}} & {\textbf{COCO}} & {\textbf{VG}} \\ \midrule
            SG2IM & 41.7 & 16.9 \\
            WSGC & 41.9 & 18.0 \\
            \textbf{Ours} & \textbf{42.0} & \textbf{20.6} \\ \bottomrule
        \end{tabular}
        \label{tab:layout-gen}
\end{table}

\label{sec:results}

\section{Ablation study}

The core components of our proposed approach rely on the multi-head attention mechanism to encode the graph nodes and to conditionally generate the image.
We perform ablation studies on the components that constitute our main contribution in this work: the SGTransformer module and the Image Transformer. 
We first evaluate the impact of edge features $E$ in the graph self-attention and the Laplacian positional encoding on our \emph{SGTransformer}. Table~\ref{tab:enc_ablation} reports the results in terms of mIoU of the predicted layout with respect to the ground truth. It shows that taking edge features into account when computing attention in the graph is crucial to predict a more structured layout, as edge features contain useful information on the type of relations among the nodes. Adding the Laplacian positional encoding contributes to enrich the knowledge encoded by the transformer with the implicit geometric information of the graph.

\begin{table}[!ht]
    \centering
    \caption{Ablation study on the SGTransformer architecture, analyzing the contribution of the self-attention using the edge features as in Equation and the Laplacian positional encoding. Results are expressed in mIoU.}
    \begin{tabular}{lrrr}
    \toprule
        \textbf{Model} &\textbf{COCO} & \textbf{CLEVR} & \textbf{VG} \\ \midrule
        SGTransformer & 21.24 & 35.48 & 9.45  \\
        \hspace{0.15cm} $+ E$ & 39.33 & 37.26 & 14.62 \\
        \hspace{0.25cm} $+ LapPE$ & \textbf{42.03} & \textbf{40.63} & \textbf{20.63}  \\ \bottomrule
    \end{tabular}
    \label{tab:enc_ablation}
\end{table}

We then evaluate the contribution of the chosen attention strategy in our \emph{Image Transformer} to correctly predict autoregressively the sequence of codebook indices, which are then used for image synthesis. 
We specifically test two architectural variants of the  \emph{Image Transformer}:
\begin{itemize}
    \item \emph{CrossAtt-ImT} that uses latent scene graph representation, learned by  our \emph{SGTransformer}, for cross-attention with the sequence predicted by the decoder. This follows the original decoder architecture proposed in~\citep{transformer}. 
    \item \emph{SelfAtt-ImT} that employs a GPT~\citep{gpt} architecture and computes only self-attention on the input sequence, using the intermediate layout (estimated by \emph{SGTransformer}) as conditioning information. This is the approach described in Sect.~\ref{sec:method}.
\end{itemize}

\begin{table}[!ht]
    \caption{Ablation study on the Image Transformer architecture. \emph{CrossAtt-ImT} uses cross-attention between the SGTransformer output and the codebook index sequence. \emph{SelfAtt-ImT}, instead,  concatenates the predicted layout with the decoder input and performs self-attention on the sequence.}
    \centering
    \begin{adjustbox}{max width=0.48\textwidth}
    \begin{tabular}{lrrrr}
        \toprule
        \multirow{2}{*}{\textbf{Model}} & \multicolumn{2}{c}{\textbf{FID ($\downarrow$)}} & \multicolumn{2}{c}{\textbf{Inception Score ($\uparrow$)}} \\ 
        \cmidrule(lr){2-3} \cmidrule(lr){4-5} 
         & COCO & CLEVR & COCO & CLEVR \\ \midrule
        \emph{CrossAtt-ImT} & 98.19 & 37.99 & 4.48 $\pm$ 0.6 & 2.57 $\pm$ 0.3 \\
        \emph{SelfAtt-ImT} & \textbf{52.34} & \textbf{32.49} & \textbf{9.82 $\pm$ 0.4} & \textbf{2.85 $\pm$ 0.7} \\ \bottomrule
    \end{tabular}
    \end{adjustbox}
    \label{tab:dec_ablation}
\end{table}

Due to the elevated computational requirements the full model requires to be trained, we evaluate a light-weight version of the models, where the SGTransformer keeps the same configuration but the number of layers in the Image Transformer is reduced to $12$, with $12$ attention heads and an embedding size of $768$.
As shown in Table~\ref{tab:dec_ablation}, \emph{SelfAtt-ImT} yields a higher image quality, showing that encoding the layout into a discrete representation and prepending it to the sequence of tokens is more effective than using the original cross-attention mechanism.

\label{sec:ablations}

\section{Conclusion}

In this paper, we propose a fully transformer-based scene graph to image approach, which exploits a multi-head attention for graph geometry learning to generate an intermediate layout representation. This representation is subsequently translated into a sequence of codebook indices by a GPT-based transformer that conditions autoregressive predictions through object location constraints. Using a lower-dimensional latent space, represented by a learned codebook of feature vectors, allows for an effective training of a high-capacity transformer model, whereas training on the pixel space would be computationally unfeasible.
Both quantitative and qualitative evaluations on standard benchmarks in the field (such as Visual Genome, COCO-Stuff, and CLEVR) show a better capability of our approach in generating high quality diverse images as well as an extensive flexibility in handling minor and major changes in the input scene graph.

\label{sec:conclusions}

\section{Acknowledgements}
We acknowledge financial support from: PNRR MUR project PE0000013-FAIR.
This work has been also supported by the research project titled ``Safe and Smart Farming
with Artificial Intelligence and Robotics'', Piano di Incentivi per la Ricerca di Ateneo
2020/2022, University of Catania (Italy).

\label{sec:acknowledgements}

\bibliographystyle{unsrtnat}
\bibliography{refs}  

\end{document}